\documentclass[conference]{IEEEtran}

\usepackage{amsmath}
\usepackage{amssymb}
\usepackage{array}
\usepackage{booktabs}
\usepackage{cite}
\usepackage{graphicx}
\usepackage{multirow}
\usepackage{tabularx}
\usepackage{url}
\usepackage[hidelinks]{hyperref}

\newcolumntype{Y}{>{\raggedright\arraybackslash}X}
\newcommand{\method}{BiRG-LoRA}
\newcommand{\acc}[1]{#1\%}
\newcommand{\resulttablewidth}{0.76\textwidth}

\title{Clinically Structured Rank-Gated LoRA for Cross-Benchmark Medical Question Answering}

\author{
\IEEEauthorblockN{Hao Gong\IEEEauthorrefmark{1},
Ruilin Gong\IEEEauthorrefmark{2},
Yining Huang\IEEEauthorrefmark{3}}
\IEEEauthorblockA{\IEEEauthorrefmark{1}
University of Chinese Academy of Sciences\\
Email: gonghao23@mails.ucas.ac.cn}
\IEEEauthorblockA{\IEEEauthorrefmark{2}
Henan University of Technology\\
Email: 251170400101@stu.haut.edu.cn}
\IEEEauthorblockA{\IEEEauthorrefmark{3}
Meta Emergence Laboratory\\
Email: huangyining1987@gmail.com}
}

\begin{document}
\maketitle

\begin{abstract}
Medical multiple-choice question answering requires parameter-efficient adaptation across heterogeneous knowledge domains and reasoning operations. A medication question, a diagnostic decision, a public-health item, and a nursing-action item may require different low-rank updates, while some recall items should preserve the base model's representation with only mild adapter intervention. We propose \method, a single-adapter rank-gated LoRA method for medical question answering. \method{} keeps one LoRA module per target layer but makes its rank dimension input-conditioned: for each question, a biaxial gate combines hidden semantic evidence with specialty/profession priors, clinical-operation priors, and their interaction to select a sparse top-$k$ subset of rank atoms. A scalar injection coefficient further controls the strength of the selected adapter update. Under a matched Qwen3-8B CMB-source protocol, \method{} achieves the highest four-benchmark macro-average accuracy among trainable PEFT baselines and matched routing controls: a \acc{69.31} average over CMB, CMExam, MedQA, and MedMCQA. It improves over MoELoRA by 0.89 percentage points while using 28.1\% fewer trainable parameters; a paired, benchmark-stratified bootstrap over final predictions gives a 95\% confidence interval of [0.42, 1.37] for this macro-average gain. Basic controls show that \method{} also improves over vanilla LoRA r16 and active-rank-matched LoRA r4 by 0.83 macro points, and an evaluation-time weak-axis perturbation check suggests that performance is not brittle to moderate tag noise. The results support a bounded claim: clinically structured rank allocation improves cross-benchmark medical QA under a matched single-seed protocol, while training-seed variance remains future work.
\end{abstract}


\section{Introduction}

Medical large language models (LLMs) are commonly evaluated with exam-style multiple-choice benchmarks such as CMB, CMExam, MedQA, and MedMCQA~\cite{wang2024cmb,liu2023cmexam,jin2020medqa,pal2022medmcqa}. These benchmarks mix heterogeneous medical specialties, professional roles, and reasoning operations. A question about drug mechanism, a treatment decision, a public-health policy item, and a nursing-action item may all share the same answer format, but they stress different regions of medical knowledge. This makes medical QA a difficult setting for ordinary parameter-efficient fine-tuning: a single fixed LoRA update can improve one subset while weakening another, whereas multi-adapter mixtures add storage, routing, and calibration complexity.

Recent LoRA routing methods address this problem by composing or selecting adapters~\cite{ostapenko2024towards,shi2024medadapter,xu2024meteora,li2025loramixer}, or by training mixture-of-LoRA experts~\cite{luo2024moelora,li2024mixlora,liao2024mingmoe,wu2024mole}. A more fine-grained direction is rank-wise routing, where the rank channels of a single LoRA are treated as small experts rather than training many complete adapters~\cite{zhao2025smora}. This is attractive for medical QA because it promises sparse activation and parameter sharing. However, a purely generic rank-wise expert view misses a crucial point: in medicine, routing signals are not arbitrary task IDs. They have meaningful structure. Specialty/profession priors, clinical-operation priors, and hidden semantic cues each describe different sources of adaptation conflict.

\begin{figure*}[t]
  \centering
  \includegraphics[width=.96\textwidth]{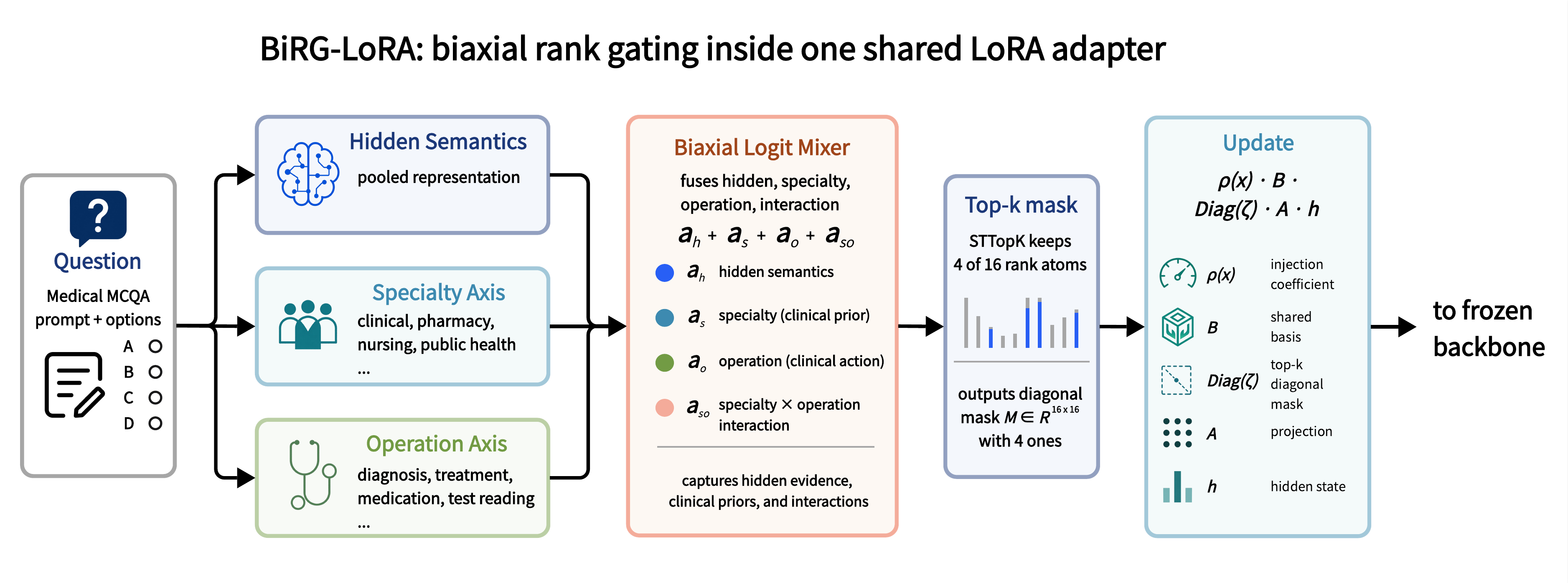}
  \caption{\method{} uses one shared LoRA basis and dynamically selects rank atoms. The gate combines hidden semantics, a specialty/profession axis, a clinical-operation axis, and their interaction. The selected top-$k$ diagonal mask and injection coefficient decide which rank atoms are active and how strongly the adapter modifies the frozen backbone.}
  \label{fig:pipeline}
\end{figure*}

We therefore ask: \emph{can a single LoRA adapter use clinically meaningful axes to allocate its internal rank channels, while remaining competitive with multi-adapter and LoRA-MoE baselines?} We propose \method, a biaxial rank-gated LoRA method. \method{} stores one rank-$R$ adapter per target module but activates only $\kappa$ rank atoms per question. The rank gate combines four signals: a hidden-state branch, a specialty/profession branch, a clinical-operation branch, and an interaction branch. A scalar injection coefficient controls the strength of the selected low-rank update. Figure~\ref{fig:pipeline} summarizes the design.

This formulation makes clinical structure part of the adapter itself. Instead of assigning every question to the same fixed rank subspace, \method{} learns which rank atoms should be used for different combinations of content domain, clinical operation, and question semantics. Under a matched Qwen3-8B protocol with 4,200 CMB training examples and 400 update steps, \method{} reaches an average accuracy of \acc{69.31} across CMB, CMExam, MedQA, and MedMCQA. It outperforms trainable MoELoRA by 0.89 points while using 66.59M rather than 92.60M trainable parameters. It also improves over vanilla LoRA controls, including rank-16 LoRA and an active-rank-matched rank-4 LoRA. Paired bootstrap tests over final predictions support the main macro-average gains, and evaluation-time axis perturbation suggests that the method is not brittle to moderate noise in the rule-derived clinical tags.

Our contributions are:
\begin{itemize}
\item We introduce a biaxial rank-gated LoRA formulation for medical QA, where specialty/profession and clinical-operation priors jointly guide sparse rank activation inside one adapter.
\item We add an input-conditioned injection coefficient that decides how strongly the selected adapter update should affect the base representation.
\item We provide a matched evaluation against base Qwen3-8B, vanilla LoRA r16/r4/r24, MoELoRA, DoRA, a generic rank-wise LoRA control inspired by SMoRA, and adapter-library baselines including MedAdapter, Arrow, MeteoRA, and LoRA-Mixer style routing.
\item We report paired, benchmark-stratified bootstrap tests and sign tests over final predictions, showing that the main macro-average gains over MoELoRA, vanilla LoRA, and a generic rank-wise control are unlikely to be explained by evaluation-set sampling noise.
\item We report ablations and robustness checks for clinical-only, hidden-only, no-injection, top-$k$, orthogonality, axis contrast, and weak-axis noise, suggesting that the full biaxial design is not reducible to generic rank-wise activation or fragile metadata lookup.
\end{itemize}

\section{Related Work}

\textbf{Medical QA and medical LLM evaluation.}
PubMedQA, MedQA, MedMCQA, CMB, and CMExam provide complementary biomedical and exam-style QA settings across English and Chinese~\cite{jin2019pubmedqa,jin2020medqa,pal2022medmcqa,wang2024cmb,liu2023cmexam}. Medical LLM studies such as Med-PaLM and Med-PaLM 2 show strong progress but also emphasize calibration, uncertainty, and safety-sensitive validation~\cite{singhal2022clinical,singhal2023expertlevel}. Our work does not propose a new benchmark; it studies how a medical adapter should allocate low-rank capacity when source and target benchmarks differ by language, specialty mix, and reasoning operation.

\textbf{Parameter-efficient adaptation.}
LoRA freezes the base weights and learns low-rank update matrices~\cite{hu2022lora}; QLoRA makes this practical for quantized LLMs~\cite{dettmers2023qlora}; and DoRA separates magnitude and direction updates for stronger low-rank adaptation~\cite{liu2024dora}. Adaptive rank methods such as AdaLoRA, DyLoRA, and IncreLoRA allocate rank budgets across modules or deployment ranks~\cite{zhang2023adalora,valipour2023dylora,zhang2023increlora}. \method{} is different because the active rank subset is chosen per medical question, and the router is explicitly conditioned on clinical axes rather than only on weight importance or a global deployment budget.

\textbf{Mixtures and routers for LoRA.}
Sparse MoE models route tokens or examples to a small subset of experts~\cite{shazeer2017moe,fedus2022switch}. LoRA-based MoE systems, including MoELoRA, MixLoRA, MING-MOE, and MoLE, train or combine multiple LoRA experts~\cite{luo2024moelora,li2024mixlora,liao2024mingmoe,wu2024mole}. Other methods reuse adapter libraries or route over generated candidates, including Arrow, MedAdapter, MeteoRA, and LoRA-Mixer~\cite{ostapenko2024towards,shi2024medadapter,xu2024meteora,li2025loramixer}. These systems motivate our baselines. The main difference is that \method{} performs structured rank selection inside a single adapter rather than selecting among multiple complete adapter modules.

\textbf{Rank-wise expert LoRA.}
SMoRA proposes that each LoRA rank can be treated as an independent expert and uses dynamic rank-wise activation for multi-task learning~\cite{zhao2025smora}. We build on the same broad direction but target a different problem. Table~\ref{tab:smora} summarizes the distinction. SMoRA addresses generic multi-task LoRA conflicts; \method{} addresses medical vertical adaptation conflicts, where the routing signal should reflect specialty, clinical operation, and hidden semantics. We also introduce an injection coefficient, because medical QA may require conservative use of base knowledge for some items and stronger adapter intervention for others.

\begin{table}[t]
\centering
\caption{Relation to prior rank-wise LoRA.}
\label{tab:smora}
\scriptsize
\setlength{\tabcolsep}{3.5pt}
\begin{tabularx}{\columnwidth}{lYY}
\toprule
Aspect & SMoRA & \method{} \\
\midrule
Core conflict & General multi-task LoRA conflict & Medical specialty and clinical-operation adaptation conflict \\
Rank semantics & Each rank is an independent expert & Rank atoms are scored by clinical-axis and hidden-semantic signals \\
Routing signal & Dynamic rank-wise activation & Medical priors plus input semantics \\
Update strength & Selects active ranks & Selects active ranks and injection strength $\rho(x)$ \\
Evaluation target & Multi-task transfer & CMB, CMExam, MedQA, and MedMCQA transfer \\
\bottomrule
\end{tabularx}
\end{table}

\section{Methodology}
\label{sec:method}

\subsection{Problem Setting}

Let $x$ be a medical multiple-choice question with options and gold answer $y$ during source training. A frozen LLM with parameters $\Theta_0$ is adapted using LoRA modules in selected projection layers. We train on a source set $D_s$ and evaluate on multiple benchmarks $D_t$. Target benchmark labels are used only for final reporting, not for training, threshold selection, or router calibration.

For a target linear layer $\ell$, standard LoRA computes
\begin{equation}
    z_{\ell}=W_{\ell}^{0}h_{\ell}+\frac{\alpha}{R}B_{\ell}A_{\ell}h_{\ell},
\end{equation}
where $W_{\ell}^{0}$ is frozen, $A_{\ell}\in\mathbb{R}^{R\times d_{\mathrm{in}}}$, and $B_{\ell}\in\mathbb{R}^{d_{\mathrm{out}}\times R}$. Standard LoRA activates all $R$ rank channels for every question. \method{} instead uses
\begin{equation}
    z_{\ell}=W_{\ell}^{0}h_{\ell}+
    \rho(x)\frac{\alpha}{R}B_{\ell}\operatorname{Diag}(\zeta(x))A_{\ell}h_{\ell},
    \label{eq:update}
\end{equation}
where $\zeta(x)\in\{0,1\}^{R}$ is a sparse rank mask with $\|\zeta(x)\|_0=\kappa$, and $\rho(x)\in[0,1]$ is an injection coefficient. In our main configuration, $R=16$ and $\kappa=4$.

Each rank channel corresponds to a rank-one atom:
\begin{equation}
    \Delta_{\ell,j}(h_{\ell}) = B_{\ell,:,j} A_{\ell,j,:}h_{\ell}.
\end{equation}
The method therefore compresses expert selection into the rank dimension of one LoRA adapter instead of selecting among many complete adapters.

\begin{table*}[t]
\centering
\caption{Design motivations in \method. Each component is tied to a medical adaptation concern and to an empirical check in Section~\ref{sec:results}.}
\label{tab:rationale}
\scriptsize
\setlength{\tabcolsep}{4pt}
\begin{tabularx}{\textwidth}{lYYY}
\toprule
Component & Medical motivation & Mechanism & Empirical check \\
\midrule
Single shared LoRA basis & Materializing many separately trained adapters is expensive when specialties or operations each need different behavior. & Keep one $A,B$ pair per target module and expose rank channels as reusable atoms. & 66.59M parameters; 5.24$\times$ compression relative to a 16-adapter rank-8 library. \\
Specialty/profession axis & Medical content differs by domain, such as pharmacy, nursing, public health, and basic biomedicine. & Add route-specific rank logits $a_s(s)$ to bias the selected rank atoms. & Clinical-only is useful but insufficient, showing that the axis carries signal but needs semantic correction. \\
Clinical-operation axis & The same knowledge domain may require different operations, such as diagnosis versus medication or test interpretation. & Add operation logits $a_o(o)$ and route-operation interactions $a_{so}(s,o)$. & Removing axis contrast lowers the average from 69.31 to 68.97. \\
Hidden semantic branch & Weak metadata can be noisy or too coarse for individual question wording. & Use pooled hidden state $h_x$ to produce text-conditioned logits $a_h(h_x)$. & Hidden-only reaches 68.82, but full biaxial gating is higher. \\
Top-$k$ diagonal mask & Medical QA should activate a compact subspace rather than all rank channels for every item. & Use straight-through top-$\kappa$ selection in $\operatorname{Diag}(\zeta(x))$. & Top-2 is too sparse, top-8 is close but below top-4. \\
Injection coefficient & Not every item should strongly overwrite base-model knowledge; some require conservative adaptation. & Multiply the sparse update by $\rho(x)\in[0,1]$. & Removing injection reduces average accuracy to 68.58. \\
Routing regularizers & Short SFT runs can collapse many questions onto the same rank atoms. & Add entropy, balance, orthogonality, and axis-contrast regularization. & Stronger balance, orthogonality-off, and axis-contrast-off ablations bound sensitivity. \\
\bottomrule
\end{tabularx}
\end{table*}

\subsection{Biaxial Rank Gate}

Medical questions differ along at least two clinically meaningful axes. The first is a specialty or profession axis $s(x)$, including clinical medicine, pharmacy, nursing, public health, medical technology, traditional Chinese medicine, and basic biomedicine. The second is a clinical-operation axis $o(x)$, including diagnosis, treatment, medication, test interpretation, mechanism, nursing action, public health, and knowledge recall. These weak labels are derived from existing metadata and deterministic keyword rules; they are used as structural priors rather than expert-certified explanations. At both training and evaluation time, $s(x)$ and $o(x)$ are derived only from the question stem, options, and publicly available non-answer metadata; gold answers and target benchmark labels are never used to construct routing features.

The rank gate produces logits
\begin{equation}
    a(x)=a_h(h_x)+a_s(s(x))+a_o(o(x))+a_{so}(s(x),o(x)),
    \label{eq:logits}
\end{equation}
where $h_x$ is a pooled hidden representation. The hidden branch lets the model depart from coarse metadata when the question text demands it. The specialty and operation branches encode clinical priors, and the interaction branch captures combinations such as pharmacy-medication versus clinical-diagnosis.

The sparse mask is formed by a straight-through top-$\kappa$ operator:
\begin{equation}
    p(x)=\operatorname{softmax}(a(x)/\tau),\quad
    \zeta(x)=\operatorname{STTopK}(p(x),\kappa).
\end{equation}
The forward pass uses a hard top-$\kappa$ mask, while gradients flow through the soft probabilities. This makes rank selection auditable while preserving differentiability.

\subsection{Injection Coefficient}

The scalar $\rho(x)$ controls update strength:
\begin{equation}
    \rho(x)=\sigma(w_{\rho}^{\top}h_x+b_{\rho}).
\end{equation}
The motivation is medical conservatism. Some questions can be answered from the base model's broad knowledge, and aggressive adapter updates may perturb a correct answer. Other questions need stronger domain-specific intervention. In the current experiments, $\rho(x)$ is helpful but not yet a mature answer-level risk calibrator: its mean is around 0.75 in the full model, and the no-injection ablation is lower on average. We therefore describe it as a dynamic update scaler rather than a clinical safety guarantee.

\subsection{Training Objective}

Training uses answer-only supervised fine-tuning on the source set. The loss is
\begin{equation}
    \mathcal{L}=
    \mathcal{L}_{\mathrm{ans}}
    +\lambda_{\mathrm{ent}}\mathcal{L}_{\mathrm{ent}}
    +\lambda_{\mathrm{bal}}\mathcal{L}_{\mathrm{bal}}
    +\lambda_{\mathrm{orth}}\mathcal{L}_{\mathrm{orth}}
    +\lambda_{\mathrm{axis}}\mathcal{L}_{\mathrm{axis}} .
    \label{eq:loss}
\end{equation}
$\mathcal{L}_{\mathrm{ent}}$ keeps routing probabilities from becoming degenerate, $\mathcal{L}_{\mathrm{bal}}$ discourages global rank collapse, $\mathcal{L}_{\mathrm{orth}}$ penalizes similarity among rank atoms, and $\mathcal{L}_{\mathrm{axis}}$ separates specialty and operation prototype rank distributions. These terms are included because short medical SFT runs can otherwise learn a superficially sparse but semantically collapsed gate.

\section{Experiments}
\label{sec:experiments}

\subsection{Datasets and Protocol}

The main experiments use the public \texttt{Qwen/Qwen3-8B} checkpoint~\cite{yang2025qwen3}. The source training set is a route-balanced CMB subset with 4,200 examples, 600 examples per major specialty/profession route, disjoint from CMB eval4149. Evaluation uses CMB eval4149, CMExam full, MedQA test700, and MedMCQA val700. CMB and CMExam are Chinese medical exams; MedQA and MedMCQA are English medical QA benchmarks. This design tests both source-domain performance and cross-benchmark transfer from Chinese training data to English questions.

All main trainable methods use the same training budget unless otherwise stated: 400 update steps, effective batch size 8, learning rate $10^{-4}$, maximum length 768, and answer-only SFT. \method{} uses LoRA rank $R=16$, active top-$\kappa=4$, alpha 32, rank temperature 0.5, rank entropy coefficient 0.01, rank-balance coefficient 0.001, orthogonality coefficient 0.1, and axis-contrast coefficient 0.5. Target modules are \texttt{q/k/v/o/gate/up/down} projections. We use greedy decoding, the same answer-letter parser for all methods, and the same prompt template within each language setting.

\subsection{Baselines}

We compare against basic and strong baselines. Basic controls include frozen Qwen3-8B, vanilla LoRA r16/alpha32, active-rank-matched LoRA r4/alpha8, and parameter-matched LoRA r24/alpha48. These rows answer whether the proposed rank gate is better than simply using a fixed low-rank adapter with similar stored rank, similar active rank, or similar trainable parameter count.

Trainable structured baselines include MoELoRA with 4 experts, top-2 routing, and rank-8 experts; DoRA with rank 24 and alpha 48; and a generic rank-wise LoRA control inspired by SMoRA. The rank-wise control uses the same rank-16/top-4 adapter but removes clinical rank routing, interaction terms, axis contrast, orthogonality, and injection gating. Adapter-library baselines include MedAdapter-style candidate verification, Arrow LoRA-weight routing, MeteoRA-style soft MoE, and LoRA-Mixer-style hard-soft routing. These routed baselines use fixed Qwen3 prompts, the same generated adapter-output pool, and source-only calibration; target labels are used only after policies are fixed.

\subsection{Statistical Testing}

Because the main margins are less than one percentage point on the four-benchmark macro average, we report paired statistical tests over final predictions. For each comparison, examples are paired by question identifier. The primary test is a benchmark-stratified paired bootstrap with 10,000 resamples: each bootstrap sample resamples examples within each benchmark and then averages the four benchmark-level deltas. This matches the paper's primary metric, the equal-weighted four-benchmark macro average. We also report a pooled paired comparison over all examples using wins/losses and a two-sided sign test. These tests address evaluation-set sampling variability for a fixed checkpoint; they do not estimate training-run variance across different seeds.

\subsection{Parameter Accounting}

\method{} has 66.59M trainable parameters. MoELoRA has 92.60M trainable parameters, so \method{} uses 28.1\% fewer trainable parameters. For scale, a 16-adapter rank-8 library would imply approximately 349.18M trainable adapter parameters; the single rank-gated adapter is therefore a 5.24$\times$ compression relative to that materialized library. The active capacity is also sparse: \method{} stores 16 rank atoms and activates 4 per question, with empirical effective rank about 3.02. MoELoRA activates two rank-8 experts per question, or roughly 16 active low-rank directions.

This accounting is intentionally conservative. MoELoRA is not an underpowered baseline: it has more trainable parameters and about four times the active low-rank directions per question. DoRA is also parameter-matched to \method{} at 66.87M trainable parameters, and LoRA r24 has 65.47M trainable parameters, close to \method{}'s 66.59M. The comparison therefore asks whether structured sparse rank allocation can improve average transfer, rather than whether a larger adapter simply wins by capacity.

\section{Results and Analysis}
\label{sec:results}

\subsection{Main Comparison}

\begin{figure}[t]
  \centering
  \includegraphics[width=\columnwidth]{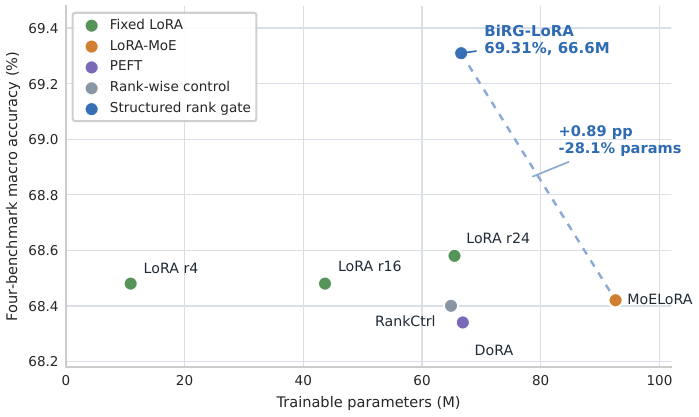}
  \caption{Accuracy--parameter Pareto view for trainable single-adapter and LoRA-MoE baselines. Markers are color-coded by method family. \method{} gives the strongest four-benchmark macro average while using fewer trainable parameters than MoELoRA.}
  \label{fig:main}
\end{figure}

\begin{figure*}[t]
  \centering
  \includegraphics[width=.80\textwidth]{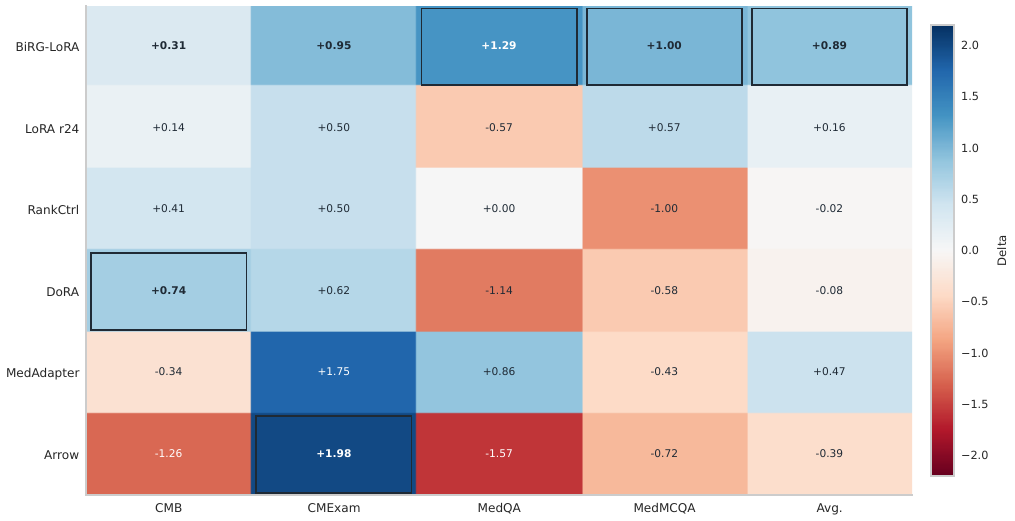}
  \caption{Benchmark-level accuracy margins relative to MoELoRA, in percentage points. Outlined cells mark the best method in each column among the displayed methods. The positive row for \method{} shows that its macro-average gain over MoELoRA is distributed across CMB, CMExam, MedQA, and MedMCQA rather than coming from a single benchmark.}
  \label{fig:margins}
\end{figure*}

Table~\ref{tab:main} reports the main Qwen3-8B CMB-source comparison. Figure~\ref{fig:main} visualizes the corresponding accuracy--parameter trade-off, and Figure~\ref{fig:margins} shows benchmark-level margins relative to MoELoRA. \method{} reaches the highest four-benchmark macro average, \acc{69.31}. It outperforms MoELoRA by 0.89 average points while using fewer trainable parameters. It also improves over vanilla LoRA r16 and active-rank-matched LoRA r4 by 0.83 average points, and over parameter-matched LoRA r24 by 0.73 average points. These basic controls are important: they show that the gain is not explained by ordinary answer-only LoRA training, by using a smaller active rank, or by matching the trainable parameter count.

The per-benchmark pattern explains why average accuracy is the primary claim. On CMB, DoRA is strongest, suggesting that a parameter-matched direction/magnitude decomposition can fit the Chinese source-like distribution very well. On CMExam, Arrow and MedAdapter are competitive because adapter-library routing preserves multiple candidate behaviors and CMExam remains close to Chinese medical exam structure. Relative to MoELoRA, however, \method{} is higher on all four benchmarks: +0.31 on CMB, +0.95 on CMExam, +1.29 on MedQA, and +1.00 on MedMCQA. The larger English-benchmark margins indicate that the sparse rank subspace learned from Chinese source data transfers to English medical questions better than the fixed-rank and full-expert alternatives in this protocol. The generic rank-wise control reaches \acc{68.40} average, below the full biaxial method, supporting the claim that rank-wise activation alone is insufficient for the medical setting.

\begin{table*}[t]
\centering
\caption{Main Qwen3-8B results under matched CMB-source training. Accuracy is reported in percent. Adapter-library rows use source-only calibration over a shared generated-output pool; N/A means the trainable-parameter count is not directly comparable to single-adapter rows.}
\label{tab:main}
\scriptsize
\setlength{\tabcolsep}{4pt}
\begin{tabular}{llrrrrrr}
\toprule
Family & Method & Params & CMB & CMExam & MedQA & MedMCQA & Avg. \\
\midrule
Base model & Qwen3-8B, no SFT & 0 & 71.68 & 68.95 & 62.86 & 54.86 & 64.59 \\
Fixed LoRA & LoRA r4/alpha8 & 10.91M & 77.83 & 75.82 & 63.00 & 57.29 & 68.48 \\
Fixed LoRA & LoRA r16/alpha32 & 43.65M & 77.87 & 76.20 & 62.43 & 57.43 & 68.48 \\
Fixed LoRA & LoRA r24/alpha48 & 65.47M & 78.09 & 75.82 & 62.57 & 57.86 & 68.58 \\
Single adapter & \textbf{\method{} r16/top4} & 66.59M & 78.26 & 76.27 & \textbf{64.43} & \textbf{58.29} & \textbf{69.31} \\
Rank-wise control & Rank-wise gate without clinical priors & 64.88M & 78.36 & 75.82 & 63.14 & 56.29 & 68.40 \\
LoRA-MoE & MoELoRA 4e/top2/r8 & 92.60M & 77.95 & 75.32 & 63.14 & 57.29 & 68.42 \\
PEFT & DoRA r24/alpha48 & 66.87M & \textbf{78.69} & 75.94 & 62.00 & 56.71 & 68.34 \\
\midrule
Adapter library & MedAdapter-style candidate verifier & N/A & 77.61 & 77.07 & 64.00 & 56.86 & 68.89 \\
Adapter library & Arrow LoRA-weight router & N/A & 76.69 & \textbf{77.30} & 61.57 & 56.57 & 68.03 \\
Adapter library & MeteoRA-style soft MoE & N/A & 76.67 & 75.33 & 61.57 & 56.57 & 67.54 \\
Adapter library & LoRA-Mixer-style hard-soft router & N/A & 77.03 & 75.52 & 62.00 & 57.14 & 67.92 \\
\bottomrule
\end{tabular}
\end{table*}

\subsection{Paired Statistical Evidence}

Table~\ref{tab:stats} reports paired bootstrap tests over final predictions. The strongest evidence is against MoELoRA, the main trainable LoRA-MoE baseline: \method{} improves the four-benchmark macro average by 0.89 points with a 95\% confidence interval of [0.42, 1.37]. The parameter-matched LoRA r24 comparison is also positive, with a macro 95\% confidence interval of [0.19, 1.28]. The comparison with the generic rank-wise control is similarly positive, supporting the role of clinical-axis-aware rank selection rather than generic rank activation alone.

The DoRA comparison is more nuanced. Since our primary metric is the equal-weighted four-benchmark macro average, the stratified bootstrap supports a macro-level advantage over DoRA. However, the pooled instance-level comparison is inconclusive because CMB and CMExam contain many more examples and DoRA is stronger on CMB. We therefore describe the DoRA result as a macro-average improvement rather than a uniformly significant per-instance gain.

\begin{table*}[t]
\centering
\caption{Paired statistical tests over final predictions. Macro CIs use benchmark-stratified paired bootstrap with 10,000 resamples. Pooled comparisons aggregate all paired examples and report a two-sided sign test over discordant pairs.}
\label{tab:stats}
\scriptsize
\setlength{\tabcolsep}{4pt}
\begin{tabular*}{\resulttablewidth}{@{\extracolsep{\fill}}lrrrrrr@{}}
\toprule
Comparison & Macro $\Delta$ & Macro 95\% CI & $P(\Delta \leq 0)$ & Pooled $\Delta$ & Pooled 95\% CI & Sign $p$ \\
\midrule
\method{} vs LoRA r16 & +0.83 & [0.33, 1.34] & 0.0007 & +0.33 & [0.01, 0.65] & 0.0462 \\
\method{} vs LoRA r24 & +0.73 & [0.19, 1.28] & 0.0046 & +0.44 & [0.08, 0.79] & 0.0179 \\
\method{} vs LoRA r4 & +0.83 & [0.28, 1.38] & 0.0014 & +0.53 & [0.19, 0.88] & 0.0026 \\
\method{} vs MoELoRA & +0.89 & [0.42, 1.37] & $<$0.0001 & +0.76 & [0.38, 1.14] & $<$0.0001 \\
\method{} vs DoRA & +0.98 & [0.39, 1.58] & 0.0007 & +0.27 & [-0.09, 0.63] & 0.1552 \\
\method{} vs rank-wise control & +0.91 & [0.43, 1.42] & 0.0001 & +0.40 & [0.11, 0.70] & 0.0085 \\
\bottomrule
\end{tabular*}
\end{table*}

\subsection{Ablation Study}

\begin{figure}[t]
  \centering
  \includegraphics[width=\columnwidth]{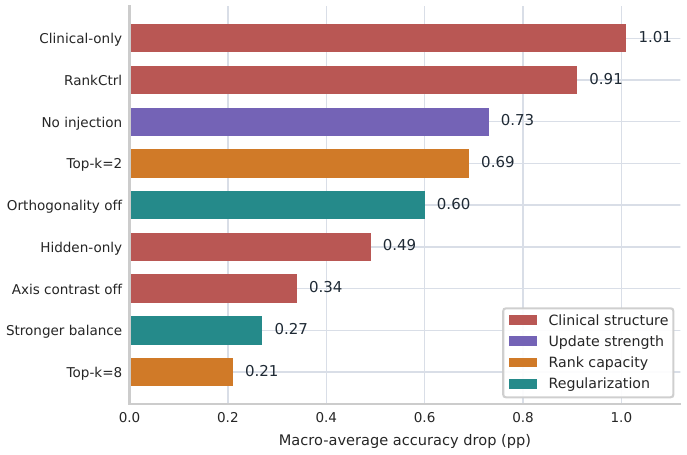}
\caption{Macro-average accuracy drop, in percentage points, relative to the full \method{} configuration. The largest drops occur when removing clinical structure, injection scaling, or adequate rank capacity.}
  \label{fig:ablation}
\end{figure}

Table~\ref{tab:ablation} and Figure~\ref{fig:ablation} summarize ablations. Clinical-only gating averages \acc{68.30}, hidden-only gating averages \acc{68.82}, and the full model reaches \acc{69.31}. The two axes are therefore complementary: clinical priors alone are too coarse, but hidden-only routing lacks the medical structure needed for robust transfer. Turning off the injection coefficient gives \acc{68.58}, supporting the usefulness of dynamic update strength. Top-$k=2$ is too sparse, while top-$k=8$ is close but lower than top-$k=4$, showing that more active rank is not automatically better.

\begin{table*}[t]
\centering
\caption{Ablation study under Qwen3-8B CMB-source training.}
\label{tab:ablation}
\scriptsize
\setlength{\tabcolsep}{3.5pt}
\begin{tabular*}{\resulttablewidth}{@{\extracolsep{\fill}}lrrrrrrr@{}}
\toprule
Variant & CMB & CMExam & MedQA & MedMCQA & Avg. & $\rho$ mean & Eff. rank \\
\midrule
Full \method{} & 78.26 & 76.27 & 64.43 & 58.29 & \textbf{69.31} & 0.749 & 3.02 \\
Stronger rank-balancing regularization & 78.19 & 76.10 & 63.86 & 58.00 & 69.04 & 0.753 & 3.05 \\
Rank-wise gate without clinical priors & 78.36 & 75.82 & 63.14 & 56.29 & 68.40 & 1.000 & 3.06 \\
Lower LoRA scaling: alpha 16 & 78.12 & 76.08 & 63.57 & 56.71 & 68.62 & 0.796 & 2.88 \\
Orthogonality off & 78.31 & 76.11 & 63.14 & 57.29 & 68.71 & 0.749 & 3.08 \\
Axis contrast off & 77.87 & 76.02 & 64.00 & 58.00 & 68.97 & 0.746 & 3.03 \\
Top-$k=2$ & 78.48 & 75.86 & 62.43 & 57.71 & 68.62 & 0.741 & 1.75 \\
Top-$k=8$ & 78.36 & 76.20 & 64.29 & 57.57 & 69.10 & 0.765 & 5.29 \\
Injection coefficient off & 78.43 & 75.60 & 62.86 & 57.43 & 68.58 & 1.000 & 3.09 \\
Clinical-only gate & 77.54 & 75.25 & 62.57 & 57.86 & 68.30 & 0.788 & 4.00 \\
Hidden-only gate & 78.33 & 75.82 & 63.71 & 57.43 & 68.82 & 0.751 & 3.01 \\
\bottomrule
\end{tabular*}
\end{table*}

\subsection{Weak-Axis Robustness}

The specialty and operation axes are weak structural priors, so a natural concern is whether the method depends on exact metadata. We therefore perturb both axes at evaluation time by randomly replacing the specialty/profession tag and the operation tag for 10\%, 20\%, or 30\% of examples using one perturbation seed. Table~\ref{tab:noise} suggests that performance is not brittle: at 30\% perturbation, the macro average is \acc{69.19}, only 0.12 points below the clean setting. This does not make the tags expert explanations, but it argues against a simple lookup-table interpretation. The hidden branch and shared rank basis appear to absorb moderate axis noise.

\begin{table}[t]
\centering
\caption{Evaluation-time perturbation of specialty/profession and clinical-operation tags using one perturbation seed.}
\label{tab:noise}
\scriptsize
\setlength{\tabcolsep}{4pt}
\begin{tabular}{lrrrrr}
\toprule
Axis noise & CMB & CMExam & MedQA & MedMCQA & Avg. \\
\midrule
0\% clean & 78.26 & 76.27 & 64.43 & 58.29 & 69.31 \\
10\% & 78.38 & 76.23 & 64.29 & 58.14 & 69.26 \\
20\% & 78.33 & 76.29 & 64.29 & 58.71 & 69.41 \\
30\% & 78.14 & 76.19 & 64.14 & 58.29 & 69.19 \\
\bottomrule
\end{tabular}
\end{table}

\subsection{Training Source and Backbone Transfer}

\begin{figure}[t]
  \centering
  \includegraphics[width=\columnwidth]{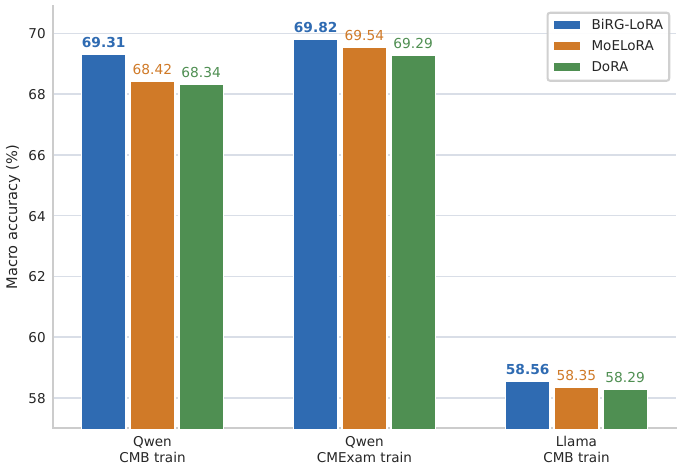}
  \caption{Macro-average accuracy across source-training and backbone settings. \method{} remains above MoELoRA and DoRA after switching the Chinese source benchmark and after replacing Qwen3-8B with Llama3.1-8B, but the Llama margin is modest.}
  \label{fig:transfer}
\end{figure}

Table~\ref{tab:transfer} reports two robustness checks, and Figure~\ref{fig:transfer} visualizes the corresponding macro-average comparisons. First, when all methods switch from CMB-source training to CMExam-source training, \method{} remains the best average method: \acc{69.82} versus \acc{69.54} for MoELoRA and \acc{69.29} for DoRA. This shows that the result is not an artifact of one Chinese training source. Second, on Llama3.1-8B-Instruct~\cite{grattafiori2024llama3}, \method{} again has the highest average, but the margin is small. We therefore treat the Llama result as supporting evidence rather than a strong standalone claim.

\begin{table}[t]
\centering
\caption{Training-source and backbone transfer. All rows use 4,200 source examples and 400 update steps.}
\label{tab:transfer}
\scriptsize
\setlength{\tabcolsep}{3.5pt}
\begin{tabular}{llrrrrr}
\toprule
Setting & Method & CMB & CMExam & MedQA & MedMCQA & Avg. \\
\midrule
\multirow{3}{*}{Qwen, CMExam train}
& \method{} & 78.21 & 77.62 & 65.00 & 58.43 & \textbf{69.82} \\
& MoELoRA & 77.97 & 77.62 & 64.14 & 58.43 & 69.54 \\
& DoRA & 78.16 & 77.55 & 63.71 & 57.71 & 69.29 \\
\midrule
\multirow{3}{*}{Llama, CMB train}
& \method{} & 56.76 & 55.63 & 62.71 & 59.14 & \textbf{58.56} \\
& MoELoRA & 55.87 & 55.97 & 61.57 & 60.00 & 58.35 \\
& DoRA & 55.75 & 54.97 & 61.86 & 60.57 & 58.29 \\
\bottomrule
\end{tabular}
\end{table}

\subsection{What the Results Support}

The experiments support a bounded claim. \method{} is not the best method on every single benchmark: DoRA is strongest on CMB in the main table, and Arrow or MedAdapter can be stronger on CMExam. The robust claim is average cross-benchmark performance under matched source training, with better parameter efficiency than MoELoRA and stronger transfer than fixed-rank LoRA and a generic rank-wise control. Paired bootstrap tests support the macro-average gains over MoELoRA, vanilla LoRA controls, and the generic rank-wise control, but they should not be interpreted as evidence of training-seed robustness.

The Chinese-to-English transfer results are also notable. Training on CMB still improves MedQA and MedMCQA relative to DoRA, MoELoRA, fixed LoRA, and the generic rank-wise control. This suggests that the rank gate is not merely memorizing Chinese benchmark tags; it learns reusable answer-domain adaptation patterns. At the same time, the Llama3.1 margins warn against overstating model independence.

\section{Limitations and Ethical Considerations}

The experiments use one training seed. We therefore do not estimate training-run variance caused by initialization, data order, or optimizer stochasticity. To reduce the risk that small margins reflect evaluation-set sampling noise, we report paired bootstrap confidence intervals and sign tests over final predictions. These tests support the main macro-average gains over MoELoRA, vanilla LoRA controls, and the generic rank-wise control, but they do not replace multi-seed training stability analysis. Small margins, especially the Llama3.1 differences, should therefore be interpreted cautiously. The evaluation is limited to multiple-choice benchmarks and does not establish clinical deployment safety.

The specialty and operation labels are weak metadata and rule-derived tags. They are useful structural priors, but they should not be read as human-validated clinical explanations. Our axis-noise check uses one perturbation seed and suggests that moderate random perturbation does not collapse performance, but it is not a substitute for expert-reviewed tags or repeated perturbation trials. Future work should test expert-curated axes and more fine-grained operation schemas.

The injection coefficient improves average accuracy, but it is not a calibrated risk score. Calibration of neural confidence is difficult~\cite{guo2017calibration}, and medical deployment would require answer-level uncertainty, verifier signals, and prospective validation. Finally, the current implementation reports active rank rather than measured latency; sparse rank computation may require fused kernels for real speedups.

\section{Conclusion}

We presented \method, a biaxial rank-gated LoRA method for medical multiple-choice QA. The method keeps one shared LoRA basis, activates a sparse subset of rank atoms per question, and conditions rank selection on hidden semantics, specialty/profession priors, clinical-operation priors, and their interaction. Under our matched Qwen3-8B source-training protocol, \method{} obtains the highest macro-average accuracy among strong PEFT, LoRA-MoE, rank-wise, vanilla LoRA, and adapter-library baselines while using fewer trainable parameters than MoELoRA. Paired bootstrap tests support the main fixed-checkpoint gains, and weak-axis perturbation suggests that the method is not fragile to moderate metadata noise. The broader lesson is that medical PEFT should ask not only how many adapter parameters to train, but which clinically structured rank subspaces should be used for each question.

\bibliographystyle{IEEEtran}
\bibliography{references}

\end{document}